Nutrition Facts, Drug Facts, and Model Facts: Putting AI Ethics into Practice in Gun Violence Research

Jessica Zhu, Dr. Michel Cukier, Dr. Joseph Richardson, Jr.

# ABSTRACT

**Objective:**

Firearm injury research necessitates using data from often-exploited vulnerable populations of Black and Brown Americans. In order to minimize distrust, this study provides a framework for establishing AI trust and transparency with the general population.

**Methods:**

We propose a Model Facts template that is easily extendable and decomposes accuracy and demographics into standardized and minimally complex values. This framework allows general users to assess the validity and biases of a model without diving into technical model documentation.

**Examples:**

We apply the Model Facts template on two previously published models, a violence risk identification model and a suicide risk prediction model. We demonstrate the ease of accessing the appropriate information when the data is structured appropriately.

**Discussion:**

The Model Facts template is limited in its current form to human based data and biases. Like nutrition facts, it also will require some educational resources for users to grasp its full utility. Human computer interaction experiments should be conducted to ensure that the interaction between user interface and model interface is as desired.

**Conclusion:**

The Model Facts label is the first framework dedicated to establishing trust with end users and general population consumers. Implementation of Model Facts into firearm injury research will provide public health practitioners and those impacted by firearm injury greater faith in the tools the research provides.

1. **OBJECTIVE**

Firearm injuries are the leading cause of death for children, adolescents and young Black men in the U.S and the annual economic toll of all firearm injuries is estimated to be 557 billion dollars. [1] Yet, firearm injury research at the Federal level, specifically the CDC and the NIH, was stymied for over two decades as result of the Dickey Amendment, until 2020 when the first Federal funding was approved since 1996. This lag in gun violence research funding has impacted the growth of machine learning (ML) used in firearm injury research and the dearth of machine learning publications on gun violence that we identified prior to 2020 – a total of three. This is in stark contrast to the number of machine publications applications in HIV research prior to 2017 alone–76. [2] Since the restrictions for federally funded gun violence research were lifted, we identified a total of eight machine learning application publications published from 2020-2023.

However, the lack of federal funding is not the only factor that has slowed machine learning (ML), and in this context, synonymously, artificial intelligence (AI), research in firearm injury. Lack of trust and transparency, coupled with very public failures in model-based policing systems have severely damaged the risk tolerance and confidence in using ML to aid in reducing firearm injuries.

Given the sensitive nature of firearm injury research and additional concerns when handling and modeling with human data, we propose the Model Facts label framework to address and mitigate concerns on AI ethics. Model Facts will provide general consumers with information they need in a digestible, unbiased format to build trust and transparency in AI systems. This framework could also be used in other domains to put AI ethics into practice. Model Facts are analogous to Nutrition Facts labels, with the addition of the warnings and usage sections from

Drug Fact labels. For instance, just as people look at Nutrition Fact labels because they may have allergies, or may be optimizing for certain nutritional macronutrients, model customers may want to search for a model with specific applications or population characteristics. With Model Facts, consumers of AI products will be able to be more deliberate in selecting the AI tools they are using without necessarily being privy to technical documentation or being completely reliant on trusting developers.

2. **BACKGROUND**

Conscious, and not just systemic biases have plagued the scientific community throughout history when working with vulnerable populations, and especially people of color. [3] The Black community continues to experience over-policing. [4] Recidivism risk prediction software has punished them more severely for nothing more than their race. [5] This has become such a concerning problem with no active solution that in 2020, coupled with protests in response to police brutality, hundreds of mathematicians boycotted collaboration with police departments, due to the risks of providing a scientific "veneer" for racism in predictive policing unless algorithms with high impact face public audit. [6] If AI tools have any hope of being applied effectively and maintaining the trust of consumers or impacted users within the firearm injury space, the AI ethics issue must be addressed.

People like to call AI a "black box". Nevertheless, just like one does not have to know exactly how a packaged food item is made in order to eat it and understand its nutritional value, an end user does not have to know the inner workings of machine learning models to be able to trust and evaluate their efficacy. As packaged food became more common, so did demand for nutritional labeling, so much so that in 1973, some level of labeling was mandated by the Federal government. [7] The modern day, standardized Nutrition Fact label was not created until 1990 with the passage of the Nutrition Labeling and Education Act. [7] Since then, they have become ubiquitous and set the template for the Drug Facts label to be published in 1999. [8] Drug Facts labels not only include components, but more importantly, uses and misuses (warnings), that consumers should take note of. Studies have shown that people are not necessarily diligent in reading these drug facts labels. [9] More than half the subjects in some studies, even when at risk of hurting themselves, admit to not reading the labels; and sometimes the labels are found to be too complex to communicate clearly with consumers. [9] However, at least these consumers know where to go to understand the contents of their product if they so choose.

Currently with AI enabled tools, consumers would likely have to go back to the original research paper or Model Card of the ML model to learn about the model and to establish trust with the system. However, there is no guarantee that the relevant information addressing biases is published and there will definitely be a lot of technical jargon for a consumer to wade through. Model Cards were first proposed in 2019 as a method to clarify a model's intended use and minimize misuse. [10] Similar to the TRIPOD checklist [11], Model Cards are a 1-2 page longform document that detail the performance metrics, features, motivation, model specifications, and ethical considerations of the model in significant depth. The authors propose that they could also be accompanied by other sheets for data.

Mitchel et al. proposed their Model Card framework to cover the breadth of knowledge and experience across ML and AI practitioners, model developers, software developers, policymakers, organizations, ML knowledgeable individuals, and impacted individuals. Since the 2019 publication, Model Cards have become commonplace in the corporate world and the open-source community, via the platform Hugging Face. Yet they primarily remain tools for developers to communicate to other technical practitioners, which is only about 50% of the original audience. In fact, as large language models have entered popular culture, many of these models and AI tools do not have Model Cards available to the public. We were not able to locate a Model Card or similar usage transparency descriptions for either ChatGPT or Bard, as of the time of writing in February 2024. OpenAI's GPT-4, one of the models behind ChatGPT, has a 60 page implementation system card that was clearly written for developers. [12] Meta's Llama 2 has a multi-page model card available on Hugging Face discussing generic training data information, general evaluation results, carbon footprint, and an ethical considerations and limitations section that can be summarized by its first and last sentences: "Llama 2 is a new technology that carries risks with use…developers should perform safety testing and tuning tailored to their specific applications of the model." [13] While end users likely would never interact with the raw Llama-2 model, nor the raw GPT-4 model, care should still be taken to ensure that these models, as the foundational components of AI systems, can be broken down and explained in a transparent fashion. They are the building blocks that provide transparency for the overall systems that tools like ChatGPT as AI tools encompass. Likewise, although Hugging Face released a Model Card API in 2022, a standardized, minimally technical, short form version still needs to be available to general public AI consumers in order to establish trust throughout the ecosystem.

There has been a significant amount of classical data science techniques applied to firearm injury research, yet there remains a gap in ML research in this field compared to other public health fields. We identified only eleven ML papers with direct impact on the field of firearm injury. [14-24] Six of these worked directly with human data, and the other five worked with human data in aggregate or trends. A nontrivial reason behind this lack of ML research can likely be attributed to the Dickey Amendment and the barriers to entry to obtain sufficient data for model training, hesitancies due to ethical complexities and Institutional Review Board (IRB) approval make it an additionally difficult area of research. In order to support collaboration between ML model developers and firearm injury practitioners, a framework must be in place to communicate findings so that policy makers, impacted individuals, and stakeholders can easily audit and trust models.

### 3. METHOD: MODEL FACTS LABEL

We propose a general utility "Model Facts" label framework to reach our target audience: end users, policy makers, organizations, and impacted individuals of AI systems or ML models. Model Facts should be separate from the Model Cards that are used by model and software developers. The Model Facts label should present information that will be easy to digest for the general population who want an overview of how well the model fits with their desired use case but are not necessarily invested in its developmental background. As such, for models that train on human datasets, like those commonly used in firearm injury prevention, we propose the following criteria and template for a generic, easily extendable Model Facts Label.

#### 3.1 Selection Criteria

Model Facts should provide enough information to readers to make a decision on whether or not the tool will fit their problem set without needing to know technical details or becoming too lengthy. As such, we prioritize information that addresses: 1) the purpose of the model, 2) the accuracy of the model, and 3) risks the model exhibits discriminatory biases. We deprioritize information on how the model was trained, how features and datasets were chosen, general dataset distributions, and categories that may be relevant to model applications but do not represent vulnerable populations. Because groups that are discriminated against change over time and bias risks may vary by application, these categories may need to be adjusted accordingly.

Like a Nutrition Facts label, a Model Facts label should only take a few minutes to read. It should not span more than one page to ensure digestibility. It also should not involve highly technical language and all metrics should also be available in a normalized setting (e.g. percentages). Systematic surveys and experiments should be conducted to evaluate the effectiveness of various features or metrics in minimizing biases and communicating model effectiveness.

**3.2 Model Facts Template**

Table 1 displays an initial recommendation of information to include in the Model Facts label. It is organized into four sections: 1) Application, 2) Accuracy, 3) Demographics, and 4) Warnings.

| **Model Facts** | |
|---|---|
| **Application:** | Brief text string describing use case |
| **Model Type:** | Classification/Regression |
| **Model Train Date:** | Date model was trained |
| **Test Data Date:** | Date range of test data |
| **Accuracy:** | **Name**        **% Over Baseline**     **Raw Score** |
|    **Optimized Score** | This is the score the model was optimized on |
|    **Standard Score** | Standard scores for given model type (Accuracy/F1/R²) |
| | **Count**          **% Train / % Test** |
| **Dataset Size:** | # of samples     % breakdown |
| **Demographics:** | **% In Test Data**    **Accuracy**        **% Target / Mean (std)** |
| **Race:** | |
|     **Asian** | |
|     **Hispanic** | |
|     **Black** | |
|     **White** | |
|     **Other** | |
| **Gender:** | This section is a breakdown of the demographics of the data used to test the model. At a minimum, developers should aim to provide information on these statistics. This table should be extendable with additional demographic groups. Accuracy is reported per the optimized score name, normalized by each group's distribution. Depending on model type, either the percent primary target of interest or the mean and standard deviation by demographic group should also be reported. |
|     **Female** | |
|     **Male** | |
|     **Trans Female** | |
|     **Trans Male** | |
|     **Nonbinary** | |
|     **Other** | |
| **Age:** | |
|     **<17** | |
|     **18-24** | |
|     **25-34** | |
|     **35-49** | |
|     **50+** | |
| **Warnings:** | Any known out of scope use cases, high risk biases, or blind spots (e.g. from untested scenarios or missing data) |

Table 1: Model Facts Template

**Application** – The first section begins with basic information about the model. The "Application" should be at most one sentence describing what the model is estimating or analyzing. In other words, this is the model's ideal use-case. "Model Type" indicates whether it is a classification or regression model, assuming this is a supervised learning tool. However, it could be extended to other model types as well, like unsupervised clustering. It may also indicate whether this is an imbalanced classification problem or any other subtypes. "Model Train Date" will tell users when the model was developed. "Test Data Dates" is important as it may be an indicator of how accurately the model can be applied on more recent evaluations.

**Accuracy** – The next section should go over the accuracy measurements of the model. This should be an easy section for end users to interpret when they are trying to choose among multiple models tested on the same application. The "Optimized Score" is the score that the model developers use to train and tune the model on while the "Standard Score" is the standardized score for that model type. For instance, for a balanced classification problem, the standard score we recommend is a standard accuracy metric: $\frac{correct\ predictions}{all\ samples}$. On the other hand, various ML models may optimize on F-1(the harmonic mean of classification precision and recall), cross-entropy, or AUC (area under the receiver operating characteristic curve) during training; these are the potential "Optimized Scores". The Names and Raw Scores are thereby only relevant when comparing models trained on the same dataset and problem. Meanwhile, the "% over Baseline" is a metric that should be the percent improvement over the base model, as defined by the model developer, and appropriately transformed if the accuracy metric is to minimize the raw score. The baseline model is often the simple approach, a computationally easier model, or sometimes assigning all variables to be in the majority class. This will show end users the value of using this model over a previous or naïve approach. It will help them conduct a cost benefit analysis of any additional biases the model may incur.

**Demographics** – Next, we breakdown the dataset. Like the "Data Nutrition Label", this will provide users with an understanding of the dataset. [25] However, our Model Facts label's dataset description will focus on a generalized understanding of only the test data by demographic groups. In our template we have highlighted race, gender, and age as these are already known to be key interacting variables in firearm injury research. Other groups and categories, like sexuality or religion, should also be included with increased data availability. Importantly, the statistics that should be reported that would allow policy makers and organizations to conduct a cursory fairness audit, are the percent of a demographic in the data, accuracy of the model in predicting that demographic, and the distribution of the target variable. The accuracy of the model in predicting that demographic should be presented as the optimized score (e.g. $\frac{correct\ predictions\ of\ female\ target\ variables}{number\ of\ female\ samples}$). If data is not collected for a certain demographic, it should be reported as such so that users know there is potential for unreported biases.

**Warnings** – Finally, the last section should include any warnings relevant to users. Like the warnings on Drug Facts, this is to inform users on when one should not trust the model in non-technical terms. For instance, if a model was trained on urban gun violence, but was later applied in rural gun violence victim predictions, this would be an out-of-scope use case. Alternatively, if a model was trained on an electively filled survey to understand community trauma, as is a common practice, there may be selection bias that is being propagated downstream. This risk should be annotated, as Lane et.al [26], do in their paper. Neither of these situations necessarily invalidates the model; it simply necessitates greater caution in validating things like potential demographics and other environmental effects prior to employing the results.

We propose that Model Facts will be useful in understanding how relevant a model is to an end user's environment and also identify where certain demographics may be over or underrepresented. While machine learning developers should already have done the appropriate trust verifications, in order to establish trust and transparency in the overall AI ecosystem, the process of filling out Model Facts will hold developers to a higher standard. We need Model Facts labels so that the general population are able to make the decision for themselves that the models are worthy of their trust.

4. EXAMPLES

In the following section we provide two examples of Model Facts labels derived from information published in peer-reviewed research.

### 4.1 Model Facts for VOID: Violent Offender Identification Directive Tool

We selected the model VOID, as it is one of the few tools actively implemented in policing. [22, 27] It is a deterministically developed model, not a ML model, though they do compare it against some ML approaches, albeit with some modeling limitations. That being said, this example is an important demonstration of how Model Facts are applicable for not just ML models. Using only the data available in the review of VOID, we constructed the Model Card in Table 2. [22]

| **Model Facts** | |
|---|---|
| **Application:** | Identify people at very high risk of near-term involvement in gun violence (suspected shooter) |
| **Model Type:** | Imbalanced Classification |
| **Model Train Date:** | 2012 |
| **Test Data Date:** | 2013 |

| **Accuracy:** | Name | % Over Baseline | Raw Score |
|---|---|---|---|
| **Optimized Score** | AUC | 10% | 0.939 |
| **Standard Score** | F1 | F1 is not reported | |

| | Count | % Train / % Test |
|---|---|---|
| **Dataset Size:** | 237232 | N/A /100% |

| **Demographics:** | % In Test Data | Accuracy | % Target / Mean (std) |
|---|---|---|---|
| **Race:** | | | |
| Asian | | | |
| Hispanic | | | |
| Black | No demographic breakdown is available in the published paper. | | |
| White | | | |
| Other | | | |
| **Gender:** | | | |
| Female | | | |
| Male | | | |
| Trans Female | | | |
| Trans Male | | | |
| Nonbinary | | | |
| Other | | | |
| **Age:** | | | |
| <17 | | | |
| 18-24 | | | |
| 25-34 | | | |
| 35-49 | | | |
| 50+ | | | |
| **Warnings:** | The probability of a high-risk individual being involved in gun violence is only around 3% when limiting to the top 1000 scores. Using prior criminal history for estimating risk may propagate any systemic policing biases. | | |

Table 2: Model Facts based on VOID's deterministic tool [22]. For demonstration purposes in this paper, we have colored cells red where no information is available.

This Model Facts Label lacks a significant amount of information. The paper clearly discusses the application and also discusses how gun violence is concentrated among a small number of individuals, which indicates this as an "imbalanced classification" model type, where a dataset has significantly fewer samples of the target category than other categories. The model train date is 2012 because that is when the police department formulated the deterministic weights used in the VOID tool; it is when the scores were last updated. The test data date is when the data used to "predict" involvement in gun violence and measure accuracy of the tool was pulled from.

We consider the "Baseline" to be the "Logit 2" model that they trained as a comparison against the VOID tool as the Logit 2 model is the worst performing model for that application. The "Optimized Score" is AUC, as that is the metric they train their baseline models to maximize. We define F-1 to be "Standard Score" because this is an imbalanced classification problem and AUC is highly sensitive to data imbalance, while F-1 corrects for the imbalance. A model that classifies all items as the majority category will have a high AUC score despite completely missing the target, minority class. The sensitivity and positive predictive value is provided which are alternate metrics for addressing accuracy in imbalanced classification problems. Nevertheless, the F-1 score for VOID is not reported. For the sake of the succinctness of the Model Facts, we believe a combined, "Standard Score", like F-1, would be better than a mix of multiple scores.

For "Dataset Size", the predictive accuracy of the VOID tool was tested on every individual in the police department's record management system who had recorded police contact between 2000 and 2012, amounting to 237,232 individuals. The risk scores for these individuals were calculated and compared to gun crime in 2013. Because the tool was constructed deterministically, albeit from data between 2009-2012, versus computationally "trained", there is no training data. Therefore, we list that 100% of the dataset went to testing.

Finally, the paper does not report any demographic characteristics of the model or the dataset in the publication. The police department in the study does not use these characteristics in their tool but does have it available. As Wheeler et al. [22] discusses, using prior criminal history likely creates a large disparity in identifying young male minorities. Publishing the demographic statistics behind the model, and making users aware of the systemic biases, will allow them to better correct for any biases after the fact. Given the workflow already involves an internal review of any model recommendations, this knowledge could easily be factored in.

### 4.2 Model Facts for Firearm Suicide Risk

For another example, we adapted the model developed in "Machine Learning Analysis of Handgun Transactions to Predict Firearm Suicide Risk" (Table 3). [20]

| Model Facts | | | |
|---|---|---|---|
| **Application:** | Predict firearm suicide risk within one year of a handgun transaction record from California | | |
| **Model Type:** | Imbalanced Classification | | |
| **Model Train Date:** | 19-May-2022 | | |
| **Test Data Dates:** | 01-Jan-1996 to 06-Oct-2015 | | |
| **Accuracy:** | Name | % Over Baseline | Raw Score |
| Optimized Score | AUC | Baseline is not defined. | 0.8 |
| Standard Score | F1 | | 0.067 |
| | Count | % Train / % Test | |
| **Dataset Size:** | 4976391 | 70% / 30% | |
| **Demographics:** | % In Test Data | Accuracy | % Target / Mean (std) |
| **Race:** | | | |
| Asian | Data is available but not broken down with these specific statistics. | | |
| Hispanic | | | |
| Black | | | |
| White | | | |
| Other | | | |
| **Gender:** | | | |
| Female | | | |
| Male | | | |
| Trans Female | Unknown if this level of demographic detail is available. | | |
| Trans Male | | | |
| Nonbinary | | | |
| Other | | | |
| **Age:** | | | |
| <17 | | | |
| 18-24 | | | |
| 25-34 | | | |
| 35-49 | | | |
| 50+ | | | |
| **Warnings:** | The suicide risk of people who commit suicide more than a year after purchasing a firearm is not modeled. Attempted suicide risk also is not modeled. | | |

Table 3: Model Facts based on predicting firearm suicide risk using handgun transactions [20]. Green cells indicate data is likely available though not accessible; yellow cells mean it is unknown; red cells indicate that no information is available.

This paper proposes a model developed to predict firearm suicide risk within one year of a handgun transaction from California. Because the authors follow the TRIPOD checklist [11], the only information missing is an accuracy comparison against a baseline model. Like the previous example, the "Model Type" also is imbalanced classification, as people who commit suicides within one year of purchasing a handgun are rare. The authors list that analysis ended May 19, 2022, so we assume this is the model train date. The dataset is from January 1, 1996 to October 6, 2015 and because they randomly sample from it to create the dataset, we assume the test data range is from this same time range. Various accuracy metrics are provided as well.

Their paper breaks down demographic information by race, gender, and age although not in the same format as what we recommend in the Model Facts template. As the researchers have this data, it would be a trivial effort for the model developers to obtain these statistics if Model Facts became the standard and their model was deployed. Since we do not have access to their raw data and they use data going back to 1996, we cannot confirm to what granularity gender identity is broken down. However, this could be updated in future versions. If this model were to be used by the California Department of Justice and directly impact potential gun buyers as Laqueur et al [20], posit, and came with a Model Facts label, retailers and buyers could access the short form key information and operate with greater trust in the models that impact their day to day lives and constitutional freedoms.

5. **DISCUSSION AND FUTURE WORK**

Model Facts should become a requirement for ML tools that use human data, especially for models that are applied to government police work and used in predicting suspects of firearm violence. Release of this information will increase trust in both the model and criminal justice system. It will also increase the ability for the corresponding tools to be audited for fairness.

Prior to becoming policy, testing must be completed to validate what formats, fields, and statistics are more valuable to end users for different model types in order to critically evaluate the fairness of these models and establish trust in AI models. For instance, experiments should explore if a second page of ancillary information, like with the back of a Drug Facts Label, is necessary or valuable to consumers. Information like model developer or data source may be useful in order to identify potential conflicts of interest. Experiments may also explore if an absolute value metric, like % Daily Value, that compares demographic information to known distributions for various applications, is useful or extendable. For example, if the application is U.S. urban firearm injury risk prediction, then the demographics will compare the model's dataset against the average demographics in overall U.S. urban environments.

For Model Facts to become truly accessible to everyday consumers, educational curricula, like Nutrition Fact Label campaigns, will also need coordination. Finally, future work should look into methods of extending this framework for models trained with datasets where risks of bias might not be overtly apparent, to include not-human data and mass datasets used in efforts like large language models. Using ethically validated models to estimate demographic information of the datasets could be a starting point for generating information on these large datasets.

## 6. CONCLUSION

We have proposed Model Facts, a novel modular template for communicating model information to general population consumers. It is designed with firearm research's commonly prevalent human based data in mind but is extendable to other data sources and applications. This data and the models resulting from firearm injury research are extremely vulnerable to misuse and biases in various demographic enclaves. As such, we offer a succinct and simple way to display basic demographic information of the model's testing environment so that interested consumers can compare the model's relevance as needed.

Ultimately if ML developers all publish Model Facts for their end users, end users may ultimately establish greater trust in AI tools and have a better understanding of how they are being represented by these tools. This will allow end users to take greater control over how and what AI tools they want to interact with. It will be beneficial for the greater ecosystem if firearm injury researchers are able to take advantage of the wealth of capabilities in machine learning research and machine learning researchers are able to communicate the limitations of their approaches effectively to their counterparts.

## 7. REFERENCES

1 Song J, Topaz M, Landau AY, et al. Using natural language processing to identify acute care patients who lack advance directives, decisional capacity, and surrogate decision makers. *PLOS ONE*, 17(7):e0270220, Jul 2022.

2 Bisaso KR, Anguzu GT, Karungi GA, et al. A survey of machine learning applications in HIV clinical research and care. *Computers in Biology and Medicine*, 91:366–371, December 2017

3 Jaiswal J. Whose responsibility is it to dismantle medical mistrust? Future directions for researchers and health care providers. *Behavioral Medicine*, 45(2):188–196, April 2019.

4 Jones-Brown D and Williams JM. Over-policing black bodies: the need for multidimensional and transformative reforms. *Journal of Ethnicity in Criminal Justice,* 19(3–4):181–187, October 2021.

5 Angwin J, Larson J, Mattu S, et al. Machine bias. *ProPublica* Online, 23 May 2016. https://www.propublica.org/article/machine-bias-risk-assessments-in-criminal-sentencing (accessed 8 Feb 2024).

6 Linder C. Why hundreds of mathematicians are boycotting predictive policing. *Popular Mechanics* Online, July 20, 2020. https://www.popularmechanics.com/science/math/a32957375/mathematicians-boycott-predictive-policing/ (accessed 8 Feb 2024).

7 Institute of Medicine (US) Committee on Examination of Front-of-Package Nutrition Rating Systems and Symbols; Wartella EA, et al. 2, History of Nutrition Labeling. In: Front-of-Package Nutrition Rating Systems and Symbols: Phase I Report. Washington (DC): National Academies Press (US); 2010. Available from: https://www.ncbi.nlm.nih.gov/books/NBK209859/.

8 FDA. Otc drug facts label. Online, June 5, 2015. https://www.fda.gov/drugs/information-consumers-and-patients-drugs/otc-drug-facts-label (accessed 8 Feb 2024).

9 Catlin J and Brass E. The effectiveness of nonprescription drug labels in the United States: Insights from recent research and opportunities for the future. *Pharmacy*, 6(4):119, October 2018.

10 Mitchell M, Wu S, Zaldivar A, et al. Model cards for model reporting. FAT* '19: Conference on Fairness, Accountability, and Transparency, January 29–31, 2019, Atlanta, GA, USA, January 2018.

11 Collins GS, Reitsma JB, Altman DG, et al. Transparent reporting of a multivariable prediction model for individual prognosis or diagnosis (TRIPOD): The TRIPOD statement. *Annals of Internal Medicine*, 162(1):55–63, January 2015.

12 OpenAI. Gpt-4 system card. Online, March 23, 2023. https://cdn.openai.com/papers/gpt-4-system-card.pdf (accessed 8 Feb 2024).

13 Meta. Llama-27b-chat-hf, Online, July 2023. https://huggingface.co/meta-llama/Llama-2-7b-chat-hf (accessed 8 Feb 2024).

14 Swedo EA, Alic A, Law RK, et al. Development of a machine learning model to estimate us firearm homicides in near real time. *JAMA Network Open*, 6(3):e233413, March 2023.

15 Polcari AM, Hoefer LE, Zakrison TL, et al. A novel machine-learning tool to identify community risk for firearm violence: The firearm violence vulnerability index. *Journal of Trauma and Acute Care Surgery*, 95(1):128–136, April 2023.

16 Rasool M and Maphosa M. Exploring global gun violence prediction through machine learning models. International Conference on Artificial Intelligence and its Applications, 2023:205–211, November 2023.

17 Chatterjee E and Chatterjee A. Pose4gun: A pose-based machine learning approach to detect small firearms from visual media. *Multimedia Tools and Applications*, September 2023.

18 Kafka JM, Moracco KE, Pence BW, et al. Intimate partner violence and suicide mortality: a cross-sectional study using machine learning and natural language processing of suicide data from 43 states. *Injury Prevention,* ip-2023–044976, October 2023.

19 Zhou W, Prater LC, Goldstein EV, et al. Identifying rare circumstances preceding female firearm suicides: Validating a large language model approach. *JMIR Mental Health*, 10:e49359, October 2023.

20 Laqueur HS, Smirniotis C, McCort C,et al Machine learning analysis of handgun transactions to predict firearm suicide risk. *JAMA Network Open*, 5(7):e2221041, July 2022.

21 Kafka JM, Fliss MD, Trangenstein PJ, et al. Detecting intimate partner violence circumstance for suicide: development and validation of a tool using natural language processing and supervised machine learning in the national violent death reporting system. *Injury Prevention*, 29(2):134–141, December 2022.

22 Wheeler AP, Worden RE, and Silver JR. The accuracy of the violent offender identification directive tool to predict future gun violence. *Criminal Justice and Behavior*, 46(5):770–788, January 2019.

23 Goin DE, Rudolph KE, and Ahern J. Predictors of firearm violence in urban communities: A machine-learning approach. *Health & Place*, 51:61–67, May 2018.

24 Wang N, Varghese B, and Donnelly PD. A machine learning analysis of twitter sentiment to the Sandy Hook shootings. In 2016 IEEE 12th International Conference on e-Science (e-Science). IEEE, October 2016.

25 Holland S, Hosny A, Newman S, et al. The Dataset Nutrition Label: A Framework to Drive Higher Data Quality Standards. *Hart Publishing*, 2020.

26 Lane SD, Rubinstein RA, Bergen-Cico D, et al. Neighborhood trauma due to violence: A multilevel analysis. *Journal of Health Care for the Poor and Underserved*, 28(1):446–462, 2017.

27 Criminal Justice Knowledge Bank. Violent offender identification directive, Albany police department. New York State. Online July 2022. https://knowledgebank.criminaljustice.ny.gov/violent-offender-identification-directive (accessed 8 FEB 2024).

**FUNDING**

University of Maryland

**CONFLICT OF INTEREST STATEMENT**

The authors report no financial conflicts of interest.

**ACKNOWLEDGEMENTS**

Thank you to my friends, mentors, and research labmates at TRAVAIL and CRR, especially Ciro Pinto-Coelho, Jordan Ray, and William Wical, for the ideas and support in many revisions.